# Improved Calibration of Near-Infrared Spectra by Using Ensembles of Neural Network Models

Abhisek Ukil, *Member, IEEE*, Jakob Bernasconi, Hubert Braendle, Henry Buijs, Sacha Bonenfant

*Abstract*--Infrared (IR) or near-infrared (NIR) spectroscopy is a method used to identify a compound or to analyze the composition of a material. Calibration of NIR spectra refers to the use of the spectra as multivariate descriptors to predict concentrations of the constituents. To build a calibration model, state-of-the-art software predominantly uses linear regression techniques. For nonlinear calibration problems, neural network-based models have proved to be an interesting alternative. In this paper, we propose a novel extension of the conventional neural network-based approach, the use of an ensemble of neural network models. The individual neural networks are obtained by resampling the available training data with bootstrapping or cross-validation techniques. The results obtained for a realistic calibration example show that the ensemble-based approach produces a significantly more accurate and robust calibration model than conventional regression methods.

*Index Terms*—Near-infrared spectrometer, spectra, NIR, FTNIR, calibration, chemometrics, neural network, ensembles, bootstrapping, cross-validation, committee.

## I. Introduction

INFRARED (IR) or near-infrared (NIR) spectroscopy is a method used to identify a compound or to analyze the composition of a material. This is done by studying the interaction of infrared light with matter. An IR/NIR spectrum refers to the absorption of the infrared light as a function of its wavelength. In IR spectroscopy, the considered frequency range is usually somewhere between 14,000 and 10 $cm^{-1}$. Note that the frequency scale applied is wavenumbers (measured in reciprocal centimeters) rather than wavelengths (measured in microns). The absorption of the material at different frequencies is measured in percent.

'Chemometrics' is the application of mathematical and statistical methods to the analysis of chemical data, e.g., multivariate calibration, signal processing/conditioning, pattern recognition, experimental design, etc [1].

In chemometrics, calibration is achieved by using the spectra as multivariate descriptors to predict the concentrations of the different constituents. In this paper, we present and analyze the use of ensembles of neural network-based calibration models. The individual models of the ensemble are realized by resampling the original training data with bootstrapping or cross-validation techniques. The ensemble model is shown to lead to a significantly improved prediction accuracy and robustness when compared with conventional calibration approaches.

The remainder of the paper is organized as follows. In section II, background information about the work is provided. This includes information about the spectrometers used, the data sample, state-of-the-art calibration methods, and neural network-based calibration models. Section III describes the concepts behind the use of an ensemble of neural networks. An example of the application of an ensemble of neural network models for calibration purposes is discussed in section IV, and section V summarizes our conclusions.

## II. Background Information

### A. Instrument

For this work, the Fourier transform near-infrared (FTNIR) spectrometers, model ABB FTLA2000-160 [2], FTPA2000-160 [2], from ABB Analytics (Bomem), Québec, Canada were used.

### B. Dataset and Sample Spectra

The data used in this work is a set of 493 samples of a cleaning solution primarily used to remove particles from the silicon surface of computer chips. The solution has two chemical components, hereafter referred to as component 1 or C1 and component 2 or C2. The concentrations of component 1 are in the range of roughly 0 to 3%, and that of component 2 from 0 to 7%, the remaining constituent being water [2]. The spectra of such solutions, with different concentrations of components 1 and 2 were collected at different temperatures using the spectrometers mentioned above. The sample spectra were measured in single beam mode [1–2], and then transformed into relative absorbance spectra with respect to a reference spectrum [2]. This is shown in Fig. 1.

Manuscript received July 15, 2008, revised January 12, 2009.
This work was supported by ABB Corporate Research.

Abhisek Ukil (corresponding author) is with ABB Corporate Research, Segelhofstrasse 1K, Baden-Daettwil, CH-5405, Switzerland (e-mail: abhisek.ukil@ch.abb.com, phone: +41 58 586 7034, fax: +41 58 586 7358).
Jakob Bernasconi is with ABB Corporate Research, Segelhofstrasse 1K, Baden, CH-5405, Switzerland (e-mail: jakob.bernasconi@ch.abb.com).
Hubert Braendle is with ABB Corporate Research, Segelhofstrasse 1K, Baden, CH-5405, Switzerland (e-mail: hubert.braendle@ch.abb.com).
Henry Buijs is with ABB Analytical, 585 Charest Boulevard East, Suite 300, Québec G1K 9H4, Canada (e-mail: henry.l.buijs@ca.abb.com).
Sacha Bonenfant is with ABB Analytical, 585 Charest Boulevard East, Québec G1K 9H4, Canada (e-mail: sacha.m.bonenfant@ca.abb.com).

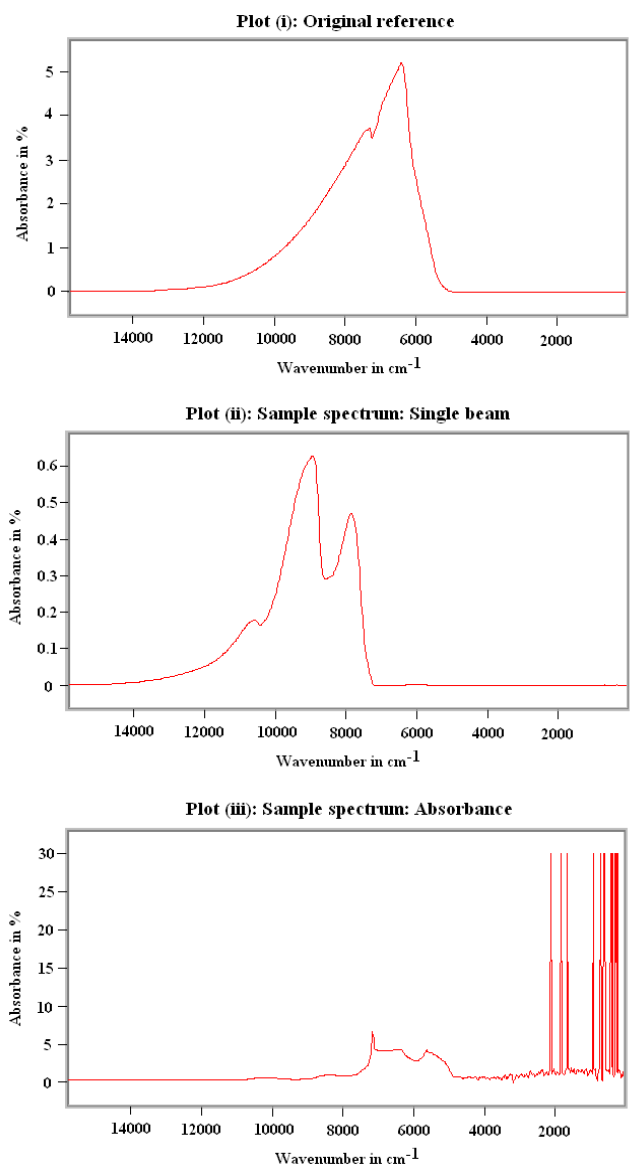

Fig. 1. Acquisition of sample spectra.

From Fig. 1, it can be seen that the region below 5000 cm$^{-1}$ cannot be used because it is below the cut-off limit of the detector, and the region between 5000–7200 cm$^{-1}$ is often saturated by water. Usually, we used the range between 7600 cm$^{-1}$ and 11000 cm$^{-1}$, where the spectra are not too noisy and the absorbance is below 1 [2].

After the acquisition, the spectra were baseline-corrected [1], and the corresponding 493 spectra, used throughout this work, are shown in Fig. 2.

*C. Calibration in Chemometrics*

The process of finding the right model parameters that lead from the spectrum to the desired information on the composition of the material is called calibration. In chemometrics, calibration is achieved by using the spectra as multivariate descriptors to predict concentrations of the constituents. The sequential steps in chemometric calibration are shown in Fig. 3. The spectra first undergo standard preprocessing techniques like multiplicative scatter correction (MSC) [1], mean centering [1], baseline correction [1], Savitzsky-Golay smoothing [3], etc to compensate for variations due to different types of instruments, changes in laboratory conditions, changes in probes, and so forth. After this, the corrected spectra are transformed into feature vectors with a reduced number of data-points. This is done using techniques such as simple wavelength selection (e.g., choosing every *n*-th wavenumber), partial least squares (PLS) [4], principal component analysis (PCA) [5], etc. A data reduction is necessary to avoid overfitting of the calibration models which would lead to a bad generalization behavior. After the preprocessing and data reduction steps, a regression model is built using the spectral feature vectors and the measured chemical concentrations from the lab-tests. The corresponding calibration model can be linear (e.g., linear regression) or nonlinear (e.g., neural networks).

State-of-the-art commercial software such as GRAMS/AI® [6], Unscrambler® [7], etc, follow the above-mentioned calibration procedure, in general by using a linear regression model.

*D. Calibration using Neural Network-based Model*

Artificial neural networks (ANNs, or simply NNs) are inspired by biological nervous systems and consist of simple processing units (artificial neurons) that are interconnected by weighted connections. The predominantly used structure is a multi-layered feed-forward network (multi-layer perceptron), i.e., the nodes (neurons) are arranged in several layers (input layer, hidden layers, output layer), and the information flow is only between adjacent layers, see Fig. 4 [8].

An artificial neuron is a very simple processing unit. It calculates the weighted sum of its inputs and passes it through a nonlinear transfer function to produce its output signal. The predominantly used transfer functions are so-called "sigmoid" or "squashing" functions that compress an infinite input range to a finite output range, e.g. [-1, +1], see [8].

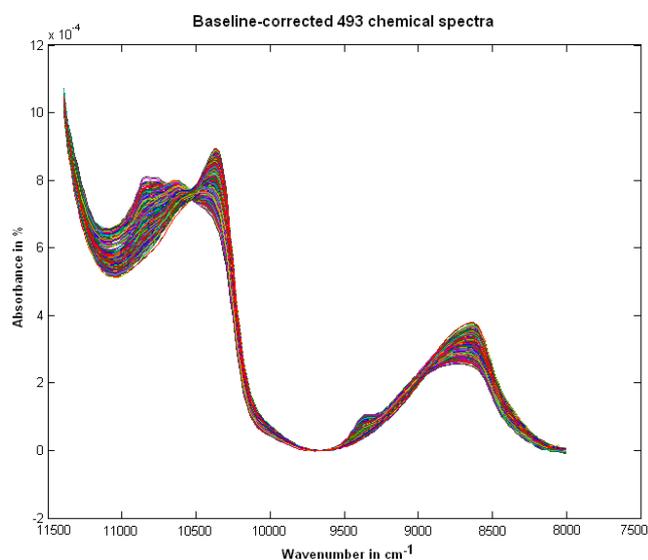

Fig. 2. The 493 baseline-corrected spectra used in this study.

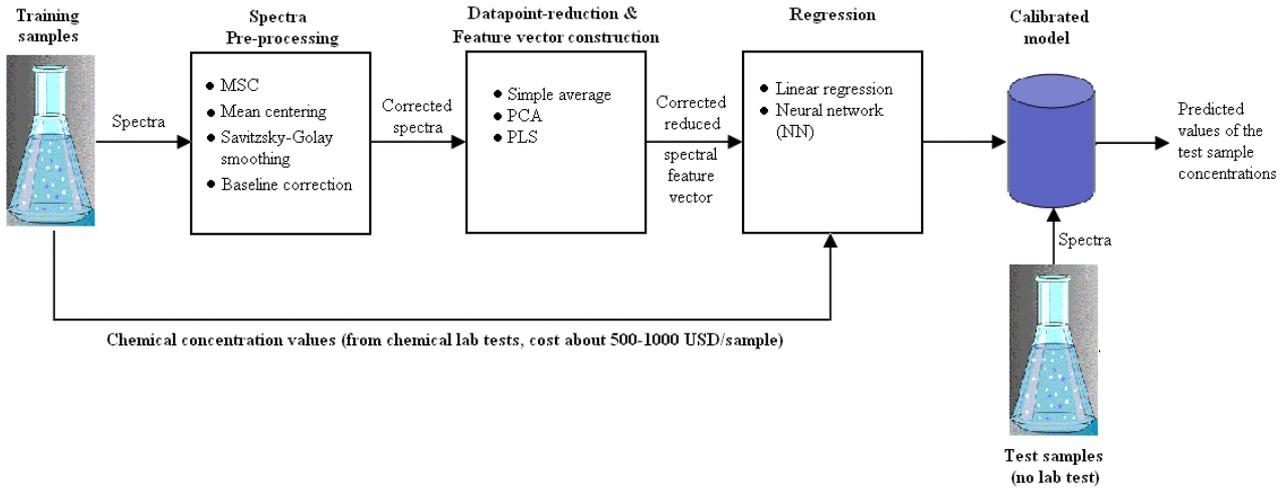

Fig. 3. Typical steps in the calibration of near-infrared spectra.

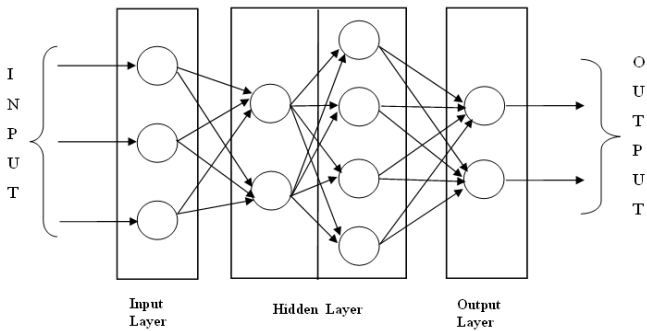

Fig. 4. Basic structure of a multi-layer perceptron.

Neural networks can be "trained" to solve problems that are difficult to solve by conventional computer algorithms. Training refers to an adjustment of the connection weights, based on a number of training examples that consist of specified inputs and corresponding target outputs. Training is an incremental process where after each presentation of a training example, the weights are adjusted to reduce the discrepancy between the network and the target output. Popular learning algorithms are variants of gradient descent (e.g., error-backpropagation) [8–9], Hebbian learning [10], radial basis function adjustments [11], etc.

Although state-of-the-art calibration software mainly relies on linear regression, neural networks are increasingly used for chemometric calibration purposes, in particular when the relationship between spectra and component concentrations is suspected to be nonlinear.

Dardenne et al. [12], e.g., discuss the use of ANN as possible candidates for multivariate calibration of NIR spectroscopic data. Geladi et al. [13] used PLS and ANN to build calibration models, and Duponchel et al. [14] used ANN for the standardization of NIR spectrometers. Benoudjit et al. [15] compared different nonlinear techniques and identified ANN-based methods as a promising technique for chemometric calibration. Kohonen neural networks [16] were applied for calibration problems in absorption spectrometry by Heyden et al. [17], and Goodacre et al. [18] applied neural network based techniques for standardization and interlaboratory calibration of mass spectrometers. Additional examples for the application of neural networks to calibration problems are cited by Brereton [1].

The motivation of our present work, however, goes beyond the use of neural networks for the calibration of NIR spectrometers. To increase the accuracy and robustness of ANN-based calibration approaches, we propose the use of an *ensemble* of neural network models. In the remainder of this paper, corresponding methodologies will be introduced, discussed, and finally applied to the data set described in section II.B.

## III. ENSEMBLES OF NEURAL NETWORK MODELS

### A. Combination of Prediction Models

Nonlinear calibration models, e.g., neural networks, have a number of advantages over conventional linear regression models. A weakness of neural networks, however, is that the corresponding learning algorithms are only guaranteed to converge to the nearest local optimum. Different initial weights, e.g., may thus lead to different calibration models. This apparent weakness may, on the other hand, be turned into a further advantage of neural network-based calibration methods.

It is well known that a combination of different prediction models can lead to a substantial improvement in prediction accuracy. In addition, corresponding ensemble predictions are also more robust than the prediction of a single model.

Here, we restrict our discussion to the simplest combination (i.e., the arithmetic average) of $n$ individual predictions $y_i(x)$, $i = 1, 2, \ldots, n$, of an unknown relation $y(x)$:

$$\bar{y}(x) = \frac{1}{n} \sum_{i=1}^{n} y_i(x). \qquad (1)$$

Using some algebra, we can then prove the following interesting relation [19]:

$$\bar{\sigma}^2(x) = \frac{1}{n}\sum_{i=1}^{n}\sigma_i^2(x) - \frac{1}{n}\sum_{i=1}^{n}(y_i(x) - \bar{y}(x))^2, \quad (2)$$

where,

$$\bar{\sigma}^2(x) = (\bar{y}(x) - y(x))^2 \quad \text{and}$$
$$\sigma_i^2(x) = (y_i(x) - y(x))^2 \quad (3)$$

denote the squared prediction errors of the ensemble and of the individual predictions respectively.

The final term on the right hand side (RHS) of (2),

$$a^2(x) = \frac{1}{n}\sum_{i=1}^{n}(y_i(x) - \bar{y}(x))^2, \quad (4)$$

is also called "ambiguity" and represents the variance of the individual predictions.

Equation (2) shows that the prediction error of the ensemble is always smaller than the average prediction error of the individual predictions. We further see that the gain in accuracy increases with increasing disagreement between the individual predictions (provided that their average prediction error does not grow proportionally).

Neural networks are particularly suited to generate different individual prediction models. We can, for example, vary the network architecture (e.g., the number of hidden units) or simply use different initial weights or different subsets of training samples.

In our analysis, we mainly used bootstrapping [21] and cross-validation [23] to generate an ensemble of different neural network-based calibration models. These two techniques are briefly explained below.

### B. Bootstrapping

The individual neural networks are trained with different training sets that are generated by randomly selecting $n$ samples from the available set of $n$ training samples. The bootstrapped training sets thus all have the same number of samples as the original training set, but in each set some of the samples are missing and some occur several times. Originally developed for estimating sampling distributions of statistical estimators from limited data, bootstrap techniques have found numerous applications in many areas of engineering [22].

### C. Cross-validation Ensembles

In a "cross-validation ensemble" of neural networks, the different training sets are generated from the available set of $n$ training samples by leaving out a given number ($m$) of samples. The training sets for the individual calibration models thus all consist of ($n$–$m$) samples, and they should preferably be chosen such that they have minimal overlap.

### D. Estimation of Confidence Intervals

It can be shown [20] that the standard deviation of the individual predictions, i.e., the square root of the ambiguity,

$$\sigma(x) = \sqrt{\frac{1}{n}\sum_{i=1}^{n}(y_i(x) - \bar{y}(x))^2}, \quad (5)$$

can be used to construct confidence intervals for the ensemble prediction,

$$\bar{y}(x) - \alpha\sigma(x) \leq y_{pred}(x) \leq \bar{y}(x) + \alpha\sigma(x), \quad (6)$$

where $\alpha$ depends on the desired confidence level.

It has to be noted, however, that these confidence intervals only reflect the modeling uncertainty for a given set of training samples and do not include prediction uncertainties that arise, e.g., from measurement errors and from different measurement conditions.

The correlation between the ensemble standard deviations, $\sigma(x)$, and the prediction errors, $\bar{\sigma}(x)$, is presented with some illustrative results from a simple test with a limited number of data (20 training and 60 test samples, taken from the dataset described in section II.B). The results shown in Fig. 5 refer to an ensemble of 10 linear regression models for component 1 (based on the first 6 principal components of the spectrum and on the temperature at which the measurement was made). The individual linear regression models correspond to different bootstrapped training sets.

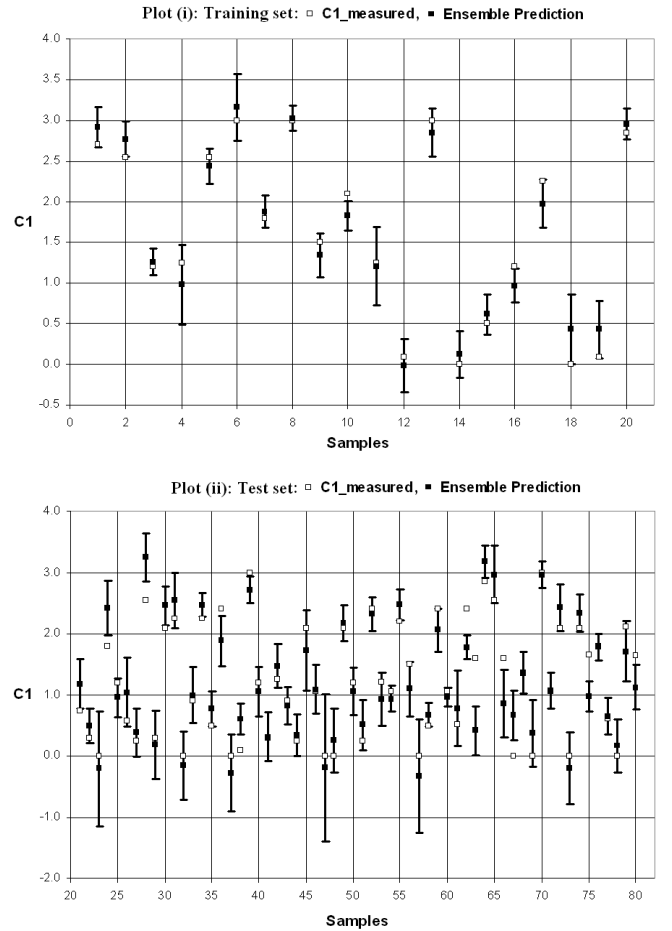

Fig. 5. Confidence intervals derived from a simple ensemble model for the calibration of spectroscopic data, plot (i) shows the 20 training samples and plot (ii), the 60 test samples from the dataset mentioned in section II.B.

The confidence intervals shown in Fig. 5 have been determined according to (6) (with $\alpha = 2$). In the training set, only one of the measured C1-values is not within the confidence interval. In the test set, however, we observe 12 of 60 samples for which the measured C1-value lies outside of the confidence interval.

IV. APPLICATION OF NEURAL NETWORK ENSEMBLES FOR CALIBRATION

In this section, we demonstrate the advantages of using an ensemble of neural network-based calibration models by applying this concept to the example dataset described in section II.B.

For the first set of test results, we started with a training set that consisted of 27 samples, randomly chosen from the available 493 samples, i.e., we had 466 samples as test data. At each iteration, we then added 30 randomly chosen samples to the training set. For the construction of feature vectors, we used PLS [4] and PCA [5] with 5 coefficients. To generate the individual models of the ensemble, we used bootstrapping (70 bootstrap models) or cross-validation (leaving out 20% of the samples). The neural network model used had 10 hidden layer nodes. To prevent overfitting, early stopping [8] (max. 20 training iterations) was used. Additionally, to overcome the problem with local minima, each model was trained 10 times with different initial weights, and the model with the best training set performance was chosen. This produced better result than averaging the 10 training runs. We also compared the results against the state-of-the-art methodologies using the GRAMS® software [6].

The chosen measure of performance is the root mean squared error (RMSE) of the test set predictions,

$$RMSE = \sqrt{\frac{1}{n}\sum_{i=1}^{n}\left(T_i^A - T_i^P\right)^2}, \quad (7)$$

where $n$ is the total number of test samples, the $T^A$s are the actually measured concentration values, and the $T^P$s are the predicted concentration values.

A. Analysis

For the different tests, we used PLS- and PCA-based feature vectors with 5 coefficients. We considered ensembles of 70 NN models that were generated with either bootstrapped or cross-validated training sets (cross-validation always referred to randomly leaving out 20% of the samples). For comparison, we also used single NN-models (with PLS feature vectors) that were trained with the same set of training samples. The RMSE-values for the test samples vs. the number of input samples for the different methods are plotted in Figs. 6 to 7, and summarized in Tables I to III. The MATLAB® neural network toolbox [24] was used for the analysis.

It is to be noted that in Figs. 6 to 7 and Tables I to III, the suffix 5 in 'PLS5' and 'PCA5' indicates number of coefficients, like five PLS or PCA coefficients. Also, in this application, no PLS regression was involved. PLS factors were used as data reduction step by using the projections of the spectra to the spectra itself (without involving any concentration), which is similar to PCA. For GRAMS® result, the optimum test predictions were considered as computed by the software. This involved usual cross-validation method to determine the optimal number of factors (latent variables) for building the model on the training set, resulting in the best test accuracy. That is why no specific number of factors are mentioned for GRAMS® result, which was different (optimally) for different test samples.

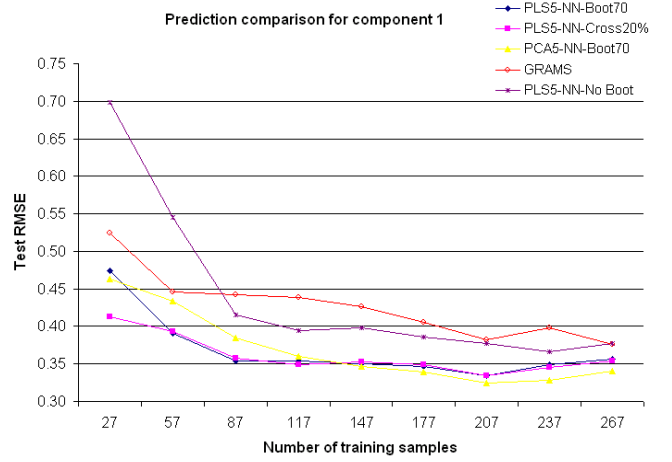

Fig. 6. Test set RMSE for component 1 vs. number of training samples: comparison of different methods.

TABLE I
TEST SET RMSE FOR COMPONENT 1: COMPARISON OF DIFFERENT METHODS

| No. of Training samples | Boot 70 PLS (5), NN | Cross 20% PLS (5), NN | Boot 70 PCA (5), NN | GRAMS | No Boot PLS (5), NN |
|---|---|---|---|---|---|
| 27 | 0.47409 | 0.41308 | 0.4629 | 0.52398 | 0.699 |
| 57 | 0.39035 | 0.39330 | 0.4334 | 0.44556 | 0.5456 |
| 87 | 0.35430 | 0.35789 | 0.3845 | 0.44258 | 0.4149 |
| 117 | 0.35421 | 0.34845 | 0.36 | 0.43801 | 0.3939 |
| 147 | 0.34977 | 0.35248 | 0.3466 | 0.42605 | 0.3985 |
| 177 | 0.34677 | 0.34902 | 0.3396 | 0.40591 | 0.3861 |
| 207 | 0.33422 | 0.33464 | 0.3243 | 0.38243 | 0.377 |
| 237 | 0.34877 | 0.34559 | 0.3287 | 0.39829 | 0.366 |
| 267 | 0.35700 | 0.35412 | 0.33990 | 0.37663 | 0.3776 |

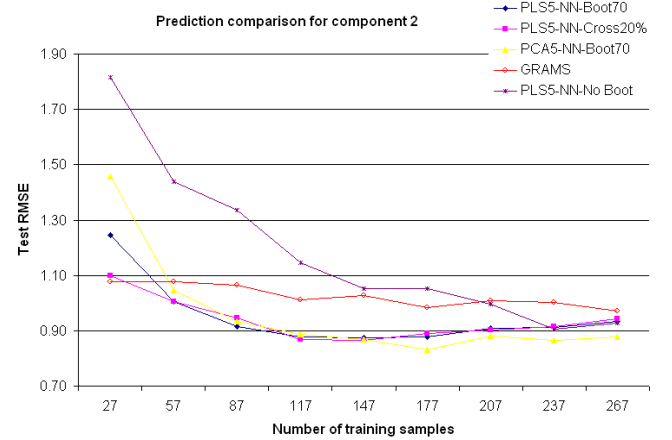

Fig. 7. Test set RMSE for component 2 vs. number of training samples: comparison of different methods.

TABLE II
TEST SET RMSE FOR COMPONENT 2: COMPARISON OF DIFFERENT METHODS

| No. of Training samples | Boot 70 PLS (5), NN | Cross 20% PLS (5), NN | Boot 70 PCA (5), NN | GRAMS | No Boot PLS (5), NN |
|---|---|---|---|---|---|
| 27 | 1.24580 | 1.10040 | 1.4581 | 1.07591 | 1.8166 |
| 57 | 1.00510 | 1.00430 | 1.0457 | 1.07845 | 1.4392 |
| 87 | 0.91460 | 0.94680 | 0.9342 | 1.06399 | 1.3356 |
| 117 | 0.87730 | 0.86960 | 0.8867 | 1.01314 | 1.1446 |
| 147 | 0.87340 | 0.86510 | 0.8669 | 1.02616 | 1.0516 |
| 177 | 0.87820 | 0.88870 | 0.8314 | 0.98384 | 1.0531 |
| 207 | 0.90730 | 0.90390 | 0.8806 | 1.00886 | 0.9951 |
| 237 | 0.91240 | 0.91520 | 0.8639 | 1.00161 | 0.9062 |
| 267 | 0.93300 | 0.94180 | 0.87660 | 0.97207 | 0.9267 |

TABLE III
ROBUSTNESS OF COMPONENT 2 PREDICTION WITH RESPECT TO CHOICE OF FEATURE VECTOR: COMPARISON OF ENSEMBLE-BASED AND SINGLE-NN APPROACH

| No. of Training samples | Boot 70 PLS (5), NN | Boot 70 PCA (5), NN | No Boot PLS (5), NN | No Boot PCA (5), NN |
|---|---|---|---|---|
| 27  | 1.24580 | 1.4581  | 1.8166 | 3.7999 |
| 57  | 1.00510 | 1.0457  | 1.4392 | 2.6431 |
| 87  | 0.91460 | 0.9342  | 1.3356 | 2.3935 |
| 117 | 0.87730 | 0.8867  | 1.1446 | 2.0514 |
| 147 | 0.87340 | 0.8669  | 1.0516 | 1.8765 |
| 177 | 0.87820 | 0.8314  | 1.0531 | 1.6721 |
| 207 | 0.90730 | 0.8806  | 0.9951 | 1.7571 |
| 237 | 0.91240 | 0.8639  | 0.9062 | 1.9252 |
| 267 | 0.93300 | 0.87660 | 0.9267 | 1.5265 |

We note that the same training and test sets were used in all different methods, i.e., we kept track of which samples were added at each iteration step and always used the same set of samples for the different methods.

*B. Discussion*

From Figs. 6 to 7 and Tables I to III, it is evident that bootstrapping and cross-validation on the training samples, i.e., the use of ensembles of neural network models generally leads to a better accuracy for predicting the concentrations of the unknown test samples than linear regression (GRAMS®) or conventional single-NN approaches.

In particular, we can make the following observations:
  i. For component 1, (Fig. 6 and Table I), we see that with a feature vector of 5 PLS coefficients, bootstrapping or cross-validation produce consistently good results. The test set accuracies for the different methods come closer when we have a relatively large number of training samples (e.g. 267 training and 226 test samples). Using a feature vector of 5 PCA coefficients and 70 bootstrapped NN-models, we achieved the best results.
  ii. The conventional NN-based approach, without involving the concept of ensembles, produces better result than GRAMS®. This substantiates the traditional usage of NN for calibration as cited in the literature (see section II.D). However, applying an ensemble of NN-models leads to a significant further improvement of prediction accuracy.
  iii. An interesting aspect is that with an ensemble of NN-based models, the number of necessary training samples to achieve a particular test set accuracy level is reduced considerably. From Table I, e.g., we see that using PLS5 and bootstrapping or cross-validation, we achieve a test set RMSE of about 0.39 with 57 training samples, while using GRAMS® [6] we need about 177 samples to achieve the same accuracy, and with a single NN-model we need about 87 samples. In comparison to the GRAMS® and single-NN approaches, we thus gain 120 and 30 samples, respectively, with an ensemble based approach. In the analytics industry, achieving a particular accuracy with less number of calibration samples is of considerable interest, as this reduces the number of laboratory tests and thus the cost of model building, see Fig. 3.
  iv. Within the ensemble-based approaches, we do not observe significant differences in the achieved test accuracies. Therefore, both bootstrapping and cross-validation techniques appear to be equally suited for generating appropriate ensembles of calibration models.
  v. For component 2 (Fig. 7 and Table II), we see that bootstrap or cross-validation ensembles with a feature vector consisting of 5 PLS coefficients produce the best results for all training set sizes.
  vi. Also in the case of component 2, the ensemble-based approaches need fewer training samples to achieve a given test set accuracy than the approaches based on a single model. From Table II, we can see that using PLS5 and bootstrapping or cross-validation, we achieve a test set RMSE of about 0.87 with 117 training samples, while with a single NN-model we need about 147 samples. With GRAMS® [6], we can never achieve such a low RMSE. The best value we have achieved is 0.97 with 267 samples.
  vii. Finally, we observe that the ensemble-based approaches appear to lead to more robust calibration models than single-NN approaches where the prediction accuracy can vary considerably for different training runs. This is illustrated in Table III, where we compare the test set accuracies of a single NN-model and of an ensemble of NN-models (for feature vectors consisting of 5 PLS or PCA coefficients, respectively). For single NN-models, we observe a large difference between the results obtained with PLS- and PCA-based feature vectors, while for the ensemble-based approaches the results are very similar. This indicates that ensemble-based approaches are much less sensitive to the choice of preprocessing methods than approaches based on a single NN-model.

V. CONCLUSION

In chemometrics, calibration of NIR spectrometers is achieved by using the spectra as multivariate descriptors to predict concentrations of the different constituents. State-of-the-art software [6–7] uses a variety of methods for preprocessing and feature vector construction, but mainly relies on linear regression techniques. However, often the relationship between spectra and component concentrations is nonlinear. Therefore, neural networks, well-known as universal nonlinear function approximator, have been used as nonlinear regression-based calibration models [12–17]. A weakness of neural networks, however, is that the corresponding learning algorithms are only guaranteed to converge to the closest local optimum. In this paper, we proposed a novel extension of conventional neural network-based calibration approaches. We introduced and discussed the use of an ensemble of neural network-based models that are generated by resampling the available calibration samples, e.g., using bootstrap or cross-validated techniques. The proposed approach was applied to a set of measured NIR spectra for a two-component cleaning solution of varying concentrations. The ensemble-based approaches showed significantly better and more robust prediction accuracy for unknown test samples when compared with models derived with state-of-the-art software or with conventional NN-based methods. Of particular interest is the observation that ensemble-based approaches can significantly reduce the number of calibration samples needed to achieve a pre-specified accuracy level.

## References

[1] R.G. Brereton, *Chemometrics: Data Analysis for the Laboratory and Chemical Plant*, John Wiley & Sons, Ltd, England, 2003.

[2] ABB Analytical, "Datasheets, user manuals, white papers, brochures," Québec, Canada. Available: http://www.abb.com/analytical

[3] A. Savitzky and M.J.E. Golay, "Smoothing and Differentiation of Data by Simplified Least Squares Procedures," *Analytical Chemistry*, vol. 36, pp. 1627–1639, 1964.

[4] I. S. Helland, "PLS Regression and Statistical Models," *Scandinavian Journal of Statistics*, vol. 17, pp. 97–114, 1990.

[5] K.H. Esbensen, D. Guyot, F. Westad, L.P. Houmoller, *Multivariate Data Analysis – In Practice*, 5th ed., Camo Process AS, Norway, 2002.

[6] Thermo Scientific, "GRAMS Spectroscopy Software Suite," 2004. Available: http://www.thermo.com/com/cda/product/detail/0,1055,22290,00.html

[7] Camo, "The Unscrambler 9.7". Available: http://unscrambler.camo.com

[8] A. Ukil, *Intelligent Systems and Signal Processing in Power Engineering*, Springer-Verlag, Heidelberg, 2007.

[9] D.E. Rumelhart, G.E. Hinton, R.J. Williams, "Learning Representations by Back-Propagating Errors," *Nature*, vol. 323, pp. 533–536, 1986.

[10] D.O. Hebb, *The Organization of Behavior*, John Wiley & Sons, New York 1949.

[11] D.S. Broomhead and D. Lowe, "Multivariable Functional Interpolation and Adaptive Networks," *Complex Systems*, vol. 2, pp. 321–355, 1988.

[12] P. Dardenne, G. Sinnaeve, V. Baeten, "Multivariate Calibration and Chemometrics for Near Infrared Spectroscopy: Which Method?," *J. Near Infrared Spectrosc.*, vol. 8, pp. 229–237, 2000.

[13] P. Geladi, H. Martens, L. Hadjiiski, P. Hopke, "A Calibration Tutorial for Spectral Data. Part 2. Partial Least Squares Regression Using Matlab and Some Neural Network Results," *J. Near Infrared Spectrosc.*, vol. 4, pp. 243–255, 1996.

[14] L. Duponchel et al., "Standardisation of Near-IR Spectrometers Using Artificial Neural Networks," *J. Molecular Structure*, vol. 480-481, pp. 551–556, 1999.

[15] N. Benoudjit et al., "Chemometric Calibration of Infrared Spectrometers: Selection and Validation of Variables by Nonlinear Models," *Chemomet. Intell. Lab. Syst.*, vol. 70, pp. 47–53, 2004.

[16] T. Kohonen, *Self-Organization and Associative Memory*, Springer-Verlag, Berlin, 1989.

[17] Y. Vander Heyden, "The Application of Kohonen Neural Networks to Diagnose Calibration Problems in Atomic Absorption Spectrometry," *Talanta*, vol. 51, pp. 455–466, 2000.

[18] R. Goodacre et al. "On Mass Spectrometer Instrument Standardization and Interlaboratory Calibration Transfer Using Neural Networks," *Analytica Chimica Acta*, vol. 348, pp. 511–532, 1997.

[19] A. Krogh and J. Vedelsby, "Neural Network Ensembles, Cross Validation, and Active Learning," In: G. Tesauro et al, eds., *Advances in Neural Information Processing Systems 7*, pp. 231–238, MIT Press, Cambridge MA, 1995.

[20] T. Heskes, "Practical Confidence and Prediction Intervals", In: M. Moser et al, eds., *Advances in Neural Information Processing Systems 9*, pp. 176–182, MIT Press, Cambridge MA, 1997.

[21] B. Efron, "Bootstrap Methods: Another Look at the Jackknife," *Ann. Statist.*, vol. 7, pp. 1–26, 1979.

[22] A.M. Zoubir and D.R. Iskander, *Bootstrap Technique for Signal Processing*, Cambridge, UK, Cambridge University Press, 2004.

[23] B. Efron, "Estimating the Error Rate of a Prediction Rule: Improvement on Cross-Validation," *J. Amer. Stat. Assoc.*, vol. 78, pp. 316–331, 1983.

[24] The MathWorks Inc., MATLAB® Neural Network Toolbox, version 7.2.0.232 (R2006a), Natick, MA, 2006.

## BIOGRAPHIES

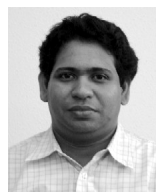

**Abhisek Ukil** (S'05, M'06) received the B.E. degree in electrical engineering from the Jadavpur Univ., Calcutta, India, in 2000 and the M.Sc. degree in electronic systems and engineering management from the Univ. of Bolton, Bolton, UK, and Southwestphalia Univ. of Applied Sciences, Soest, Germany in 2004. He was a DAAD scholar during M.Sc. He received his Ph.D. from the Tshwane Univ. of Tech., Pretoria, South Africa in 2006.

Since 2006, he is a research scientist at the 'Integrated Sensor Systems group,' ABB Corporate Research Center in Baden, Switzerland. He has published over 25 scientific papers, a monograph *Intelligent Systems and Signal Processing in Power Engineering* by Springer, Heidelberg in 2007, book chapter in *Electric Power Research Trends* by Nova Publishers, New York in 2008. His research interests include signal processing, machine learning, power systems, embedded systems.

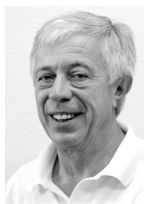

**Jakob Bernasconi** received his diploma and Ph.D degrees in theoretical physics from the ETH Zurich, Switzerland in 1968 and 1972, respectively. He was a research fellow and is currently consultant at the ABB Corporate Research Center, Baden, Switzerland. He has published over 80 scientific papers, and his research interests include disordered systems, stochastic processes, neural networks, machine learning, and economic modeling.

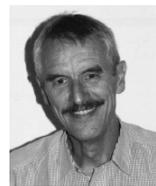

**Hubert Braendle** received the Ph.D. degree in physics from the University of Zurich, Zurich, Switzerland, in 1974.

From 1974 to 1978, he was as a Postdoctoral Researcher in high-energy physics with the University of California, Los Angeles and at the Swiss Federal Institute of Technology, Zurich, Switzerland. In 1978, he joined BBC (now ABB Switzerland, Ltd.), Baden, Switzerland. He is a research fellow and conducts research in sensor technologies, optical sensing and analytics at the 'Sensor Technologies group' at the ABB Corporate Research Center.

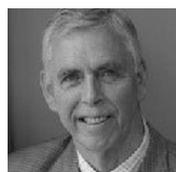

**Henry Buijs** received the BASc degree in engineering physics and the M.Sc. degree in physics from University of Toronto, Toronto, Canada. He received the Ph.D. degree in physics from the University of the British Colombia, Vancouver, Canada in 1969.

During the seventies, after having worked as a postdoctoral researcher and professor at Laval University, he along with two associates founded Bomem to commercialize the results of his research in Fourier transform infrared spectrometry, which is at the very heart of quality control processes in fields like dairy processing, pulp and paper, petrochemical industry, pharmaceutical industry and semiconductors. Currently, he is the senior technical director of ABB Analytical (Bomem) Inc., Quebec, Canada. ABB-Bomem is considered a world authority in space based spectrometry.

He received the Barringer Research Award from The Spectroscopy Society of Canada in 1978, and the Williams-Wright Award from the Coblenz society in 1998. In 2005, he received the Lionel-Boulet prize, Quebec's highest distinction in the field of industrial research and development, by Quebec's Minister for Economic Development, Innovation and Exportation.

**Sacha Bonenfant** is currently with ABB Analytical (Bomem) Inc., Quebec, Canada. He is a senior applications scientist in the spectroscopy and chemometrics field.